\documentclass{article}
\usepackage{spconf,amsmath,graphicx}
\usepackage{multirow, booktabs}
\usepackage{amssymb}

\title{Multi-layer Content Interaction through Quaternion Product for Visual Question Answering}

\name{\hspace{-10pt}
Lei Shi$^{\dagger}$\!, Shijie Geng$^{\diamond}$, Kai Shuang$^{\dagger}$ ,Chiori Hori$^{*}$, Songxiang Liu$^{\ddagger}$, Peng Gao$^{\ddagger}$, Sen Su$^{\dagger}$}

\address{\hspace{-10pt}
$^{\dagger}$Beijing University of Posts and Telecomminications \qquad $^{\diamond}$ Rutgers University
\\ $^{\ddagger}$CUHK \qquad
$^{*}$Mitsubishi Electric Research Laboratories (MERL)}

\begin{document}
\ninept
\maketitle

\begin{abstract}
Multi-modality fusion technologies have greatly improved the performance of neural network-based Video Description/Caption, Visual Question Answering (VQA) and Audio Visual Scene-aware Dialog (AVSD) over the recent years. Most previous approaches only explore the last layers of multiple layer feature fusion while omitting the importance of intermediate layers. To solve the issue for the intermediate layers, we propose an efficient Quaternion Block Network (QBN) to learn interaction not only for the last layer but also for all intermediate layers simultaneously. In our proposed QBN, we use the holistic text features to guide the update of visual features. In the meantime, Hamilton quaternion products can efficiently perform information flow from higher layers to lower layers for both visual and text modalities. The evaluation results show our QBN improve the performance on VQA 2.0,  furthermore  surpasses the approach using large scale BERT or visual BERT pre-trained models. Extensive ablation study has been carried out to examine the influence of each proposed module in this study.

\end{abstract}
\section{Introduction}
\label{sec:intro}
Multi-modality fusion technologies have been applied to neural network-based Video Description/Caption \cite{hori2017attention}, Visual Question Answering (VQA) \cite{gao2018question} and Audio Visual Scene-aware Dialog (AVSD) \cite{hori2019end} over the recent years. 
Especially, VQA connecting vision and language becomes a standard benchmark for testing fundamental multimodality fusion algorithm. 
Multi-layer Intra- and Inter- layer network has been commonly applied for multimodality feature fusion on VQA. However, most approaches including Dynamic Fusion with Intra and Inter Modality Attention Flow (DFAF)~\cite{Gao_2019_CVPR} and Modular Co-attention Network (MCAN)~\cite{yu2019deep} only uses the last layer for down-stream tasks such as VQA classification. We argue that the interaction captured in the last layer is not enough to model multimodality fusion. In fact, interaction in the intermediate layer can capture some important aspects of the interplay between intra- and inter- modality as shown in the final visualization part. 

To address the issue of missing interaction information in intermediate layers, we propose a novel Quaternion Block Networks (QBN) which can capture multi-layer interaction with a novel Hamilton quaternion product. Our proposed QBN composes every intermediate four layers as a quaternion and performs a quaternion Hamilton product between two modalities visual and language quaternions (Visual and language). The complex interaction between all layers can be captured efficiently thorough our proposed approaches.

The key contributions of the paper can be outlined
Track changes is on
 as three-fold. (1) We propose Quaternion Block Networks(QBN) to explore the impact of multi-layer content and multi-layer relationship for visual question answering. (2) We carry out spatial scaling of visual features conditional on question feature in VQA, while the existing VQA model neglects discussion on this. (3) We execute extensive ablation studies for each component of QBN and achieve state-of-the-art performance on VQA v2.0~\cite{goyal2017making}. Surprisingly, our proposed QBN can even surpass BERT retrained models like VilBERT.


\begin{figure}[t!]
 \centering
 \includegraphics[width=1.0\columnwidth]{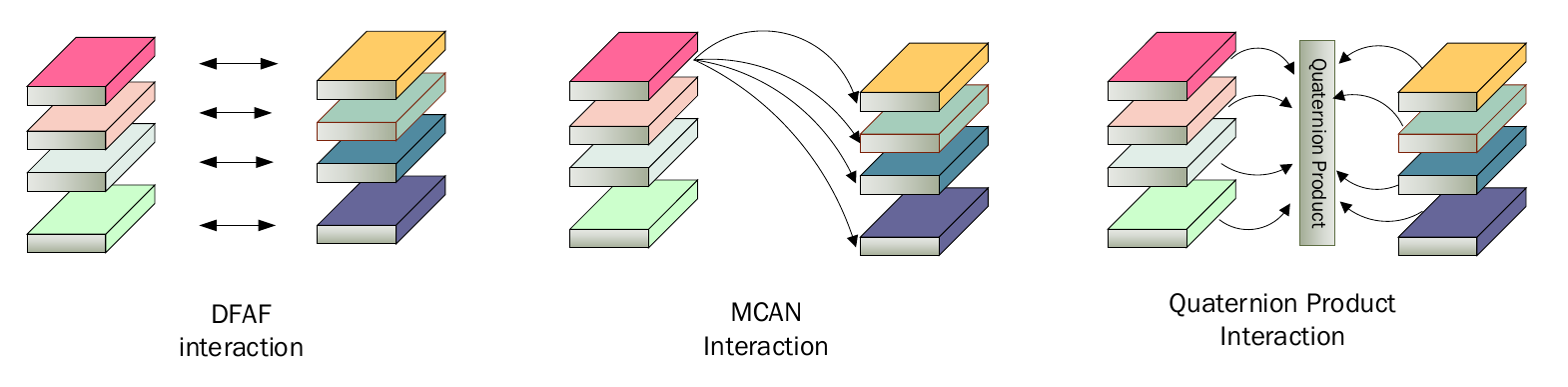}
 \vspace{-9pt}
 \caption{Three ways of multimodal information interaction. DFAF updates information between the same hierarchy of multimodal layers; MCAN only utilizes one high-level modality to update all layers in another modality; Our QBN enables updates among all layers both intra- and inter- two modalities and builds better information flow for multiple layers.}
\label{fig:compare}
\vspace{-9pt}
\end{figure}


\begin{figure*}[t!]
 \centering
 \includegraphics[width=2\columnwidth]{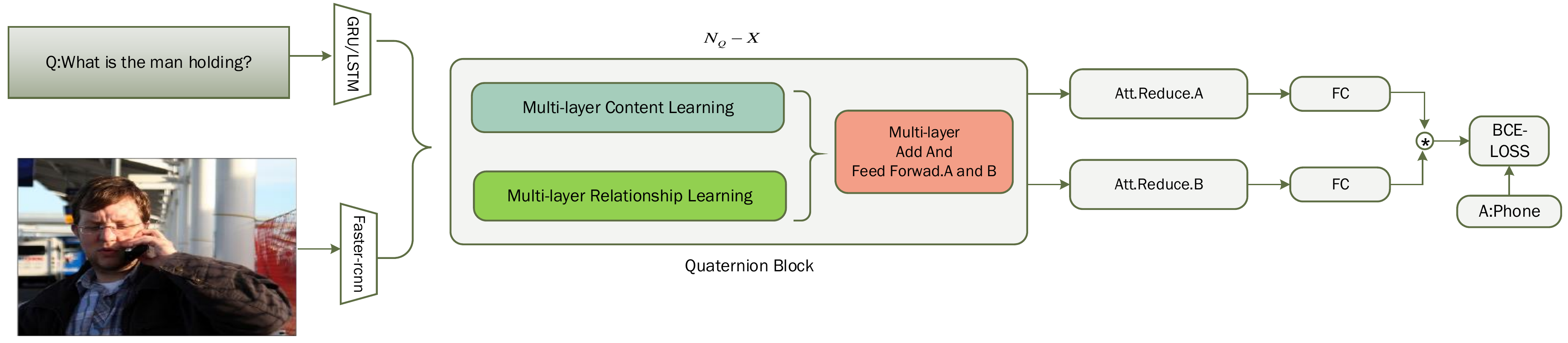}
  \vspace{-6pt}
 \caption{The overall framework of Quaternion Block Networks. We only show one Quaternion block here, the $X$ in $N_Q-X$ is the number of blocks. Each block is composed of three parts: 1) \textit{Multi-layer Content Learning module}, which learns the holistic text features to guide the update of visual features. 2) \textit{Multi-layer Relationship Learning module}, which builds connections between multi-layers of the two modalities. 3) \textit{Feed Forward module}, which integrates information of multiple layers and outputs the fused information to next Quaternion block.}
\label{fig:model}
 \vspace{-6pt}
\end{figure*}


\section{Related Work}
\label{sec:format}

\subsection{Representation Learning}
Representation learning has revolutionized both Computer Vision(CV) and Natural Language Processing(NLP). In CV, VGGNet~\cite{simonyan2014very}, ResNet~\cite{he2016deep}, SENet~\cite{hu2018squeeze} and DenseNet~\cite{huang2017densely} raise up to be the fundamental building block for downstream tasks. While in NLP, word2vec~\cite{mikolov2013distributed}, GloVe~\cite{pennington2014glove}, Skipthough~\cite{kiros2015skip}, ELMo~\cite{peters2018deep}, and BERT~\cite{devlin2018bert} play a fundamental role in pushing the research boundary. In Visual Question Answering, BERT and Faster RCNN has been treated as a fundamental feature extractor.

\subsection{Self-attention \& Co-attention}
\label{sec:pagestyle}
Attention mechanism imitates human vision by putting large weight on important regions. Neural Attention has been popularized by ~\cite{mnih2014recurrent}.  Subsequently, ~\cite{bahdanau2014neural} used the attention mechanism to perform machine translation task. Motivated by attention model in machine translation,  \cite{vaswani2017attention} proposed self-attention mechanism, which has been a fundamental backbone for many CV and NLP applications.

After its success on pure CV and NLP tasks, researchers~\cite{Gao_2019_CVPR}~\cite{yu2019deep}~\cite{Gao_2019_ICCV} start to apply self-attention and co-attention mechanism for solving multi-modality fusion among which Visual Question Answering has been served as a standard benchmark. DFAF~\cite{Gao_2019_CVPR} firstly introduce a intra- and inter- modality attention flow pipeline for approaching VQA. Separate intra-attention will be applied for message passing inside text and image, then inter-modality attention will enable cross-modality interaction. MCAN~\cite{yu2019deep} performs intra modality attention several times followed by inter modality attention while DFAF interleaves Intra and Inter modality attention. MCAN can achieve slightly better performance than DFAF. MLIN~\cite{Gao_2019_ICCV} proposes a latent interaction mechanism which can reduce the quadratic computation of self attention into linear.

\subsection{Quaternion Product}
\label{sec:typestyle}
Quaternion connectionist architectures have been widely applied in various areas such as speech recognition \cite{pavllo2018quaternet}, computer vision~\cite{parcollet2018quaternion}, and natural language processing~\cite{tay2019lightweight}. Our proposed QBN will utilize Hamilton product to learn multi-layer interaction efficiently which result in improved robustness and better interpretability.

\section{Quaternion Block Networks}
\label{sec:majhead}

\subsection{Overview}
\label{ssec:subhead}

The proposed method includes a series of Quaternion Product Block. The whole pipeline is demonstrated in Fig.2, where visual and language features from the two modalities are first pre-processed. In the second step, the two modalities consisting of the pre-processed visual and language features are serially encoded three times by the self-attention mechanism; In this way, we obtain four layers of content information in visual and text modalities respectively. In the third step,
Multi-layer textual information will be a better guide multi-modal information interaction at the appropriate level using our proposing multi-layer content operation as described in the Section \ref{subsection:multi-content}
to learn the overall presentation of multiple layers of content in text modality.
In the fourth step, 
we apply Quaternion Product mechanism in the Section \ref{subsection:multi-ship}  performing  multi-layer relationship learning 
to better learn the relationship between text modality and visual modality in multiple layers of content. 

Finally, we use the co-attention mechanism to implement multi-modal information to achieve information updates in the corresponding layer, taking into account the overall content of the multi-layer and the overall relationship of the multi-layer. Such Quaternion Product Block can be stacked multiple times to pass information iteratively flows between multi-layer words and regions to model potential alignments for visual problem-solving.
 
\subsection{Visual and language feature preprocess}
\label{sssec:subsubhead}

\begin{figure}[t!]
 \centering
 \includegraphics[width=0.9\columnwidth]{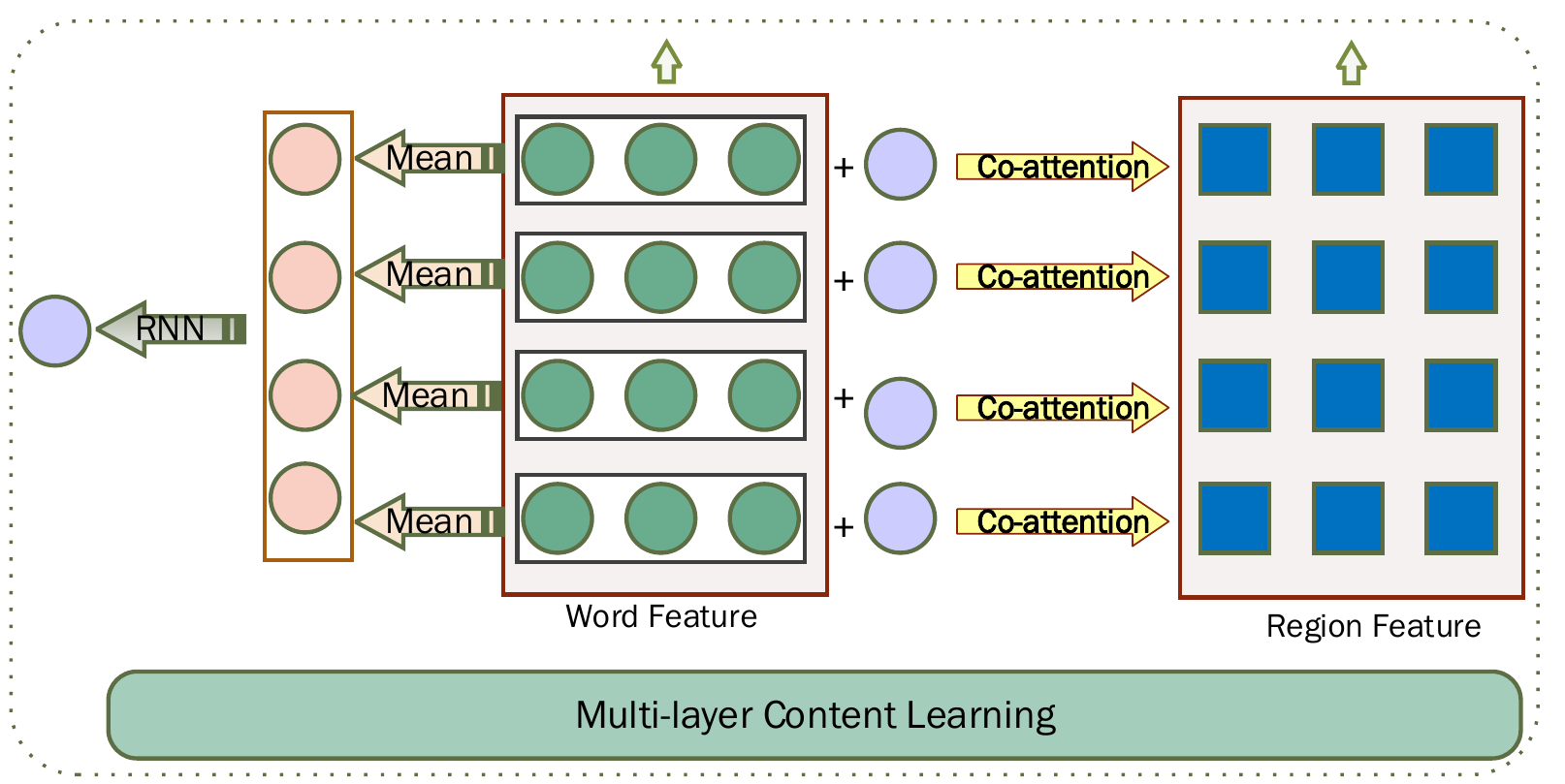}
 \vspace{-6pt}
 \caption{Illustration of multi-layer content learning in a Quaternion block. In a Quaternion block, there are four layers of word features. We first average the words features of each layer, then input the average sentence vector of each layer's word vectors into an RNN model, and then take the output hidden vector as the holistic textual feature of the whole block. Finally, we use co-attention to update the visual information of each layer in the quaternion block based on the holistic textual feature.}
\label{fig:interact}
\vspace{-6pt}
\end{figure}

To obtain basic language features, we first embed each word in question $Q$ using pre-trained Glove~\cite{pennington2014glove}. We then use an LSTM, generating a sequence of the hidden state$\left\{ {{w_1},...{w_t}} \right\}$. Note that we use the last hidden state of LSTM $w_t$ as a question feature, signified as ${q_t} \in {\mathbb{R}^{^{512}}}$and all questions are padded and truncated to the same length 14.

\begin{equation} 
    w = \mathrm{LSTM}(Q), q_t = w_t
\end{equation}%

We extract image features, which are obtained from a Faster-RCNN~\cite{ren2015faster} model pre-trained on Visual Genome~\cite{krishna2017visual} dataset. For each image I, we extract 100 region proposals and associated region features. However, different from bottom-up \& top-down attention~\cite{anderson2018bottom}, We select the $R \in {\mathbb{R}^{u\times2\times2\times2048}}$ image region feature as input. We map the dynamically changing question vector to the scaling factor and bias term of the channel feature through the fully connected layer ${f_c}$ and ${h_c}$ . The image region features are preprocessed by the sentence vector of Question to achieve effective scaling about image spatial-region features. 

\begin{equation}
     R = \mathrm{RCNN}(I)
\end{equation}
\begin{equation}
     \gamma_{i,c} = {f_c}({q_t})
\end{equation}
\begin{equation}
     \beta_{i,c} = {h_c}({q_t})
\end{equation}
\begin{equation}
     v = \mathrm{AveragePooling}\left( {{\gamma _{i,c}} * {R_{i,c}} + {\beta _{i,c}}} \right)
\end{equation}

In the formula, $\gamma $ and $\beta $ represent the scaling factor and the bias term respectively; ${q_t}$  represents question feature. In ${R_{i,c}} \in {\mathbb{R}^{2\times2\times2048}}$, $i$ represents the $i$-th region feature; $c$ represents the $c$-th channel of the 2048 channel. In $R=\{R_1,\dots,R_{\mu} \}$, $\mu$ represents an image with a total of $\mu$ region features. Finally, we average pooling the scaled region features.

\begin{figure}[t!]
 \centering
 \includegraphics[width=0.85\columnwidth]{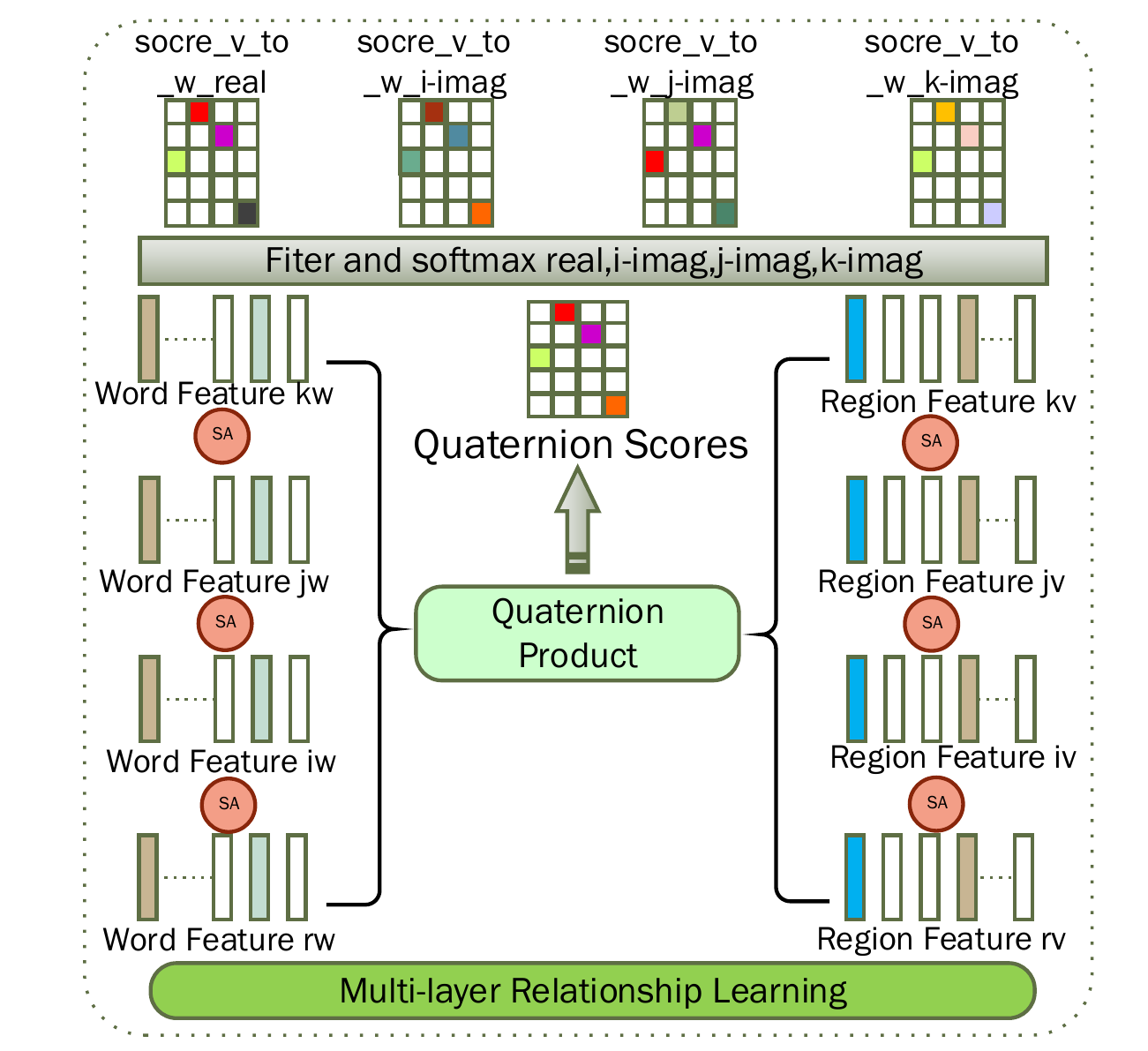}
 \caption{Illustration of multi-layer relationship learning in a Quaternion block. In a Quaternion block, we regard the textual and visual information as different parts in two Quaternions. Then we utilize Quaternion Product to conduct relationship learning among different layers across the two modalities. 
 }
\label{fig:relationship}
\end{figure}

\subsection{Multi-layer content learning for text information}\label{subsection:multi-content}
In the previous use of the self-attention~\cite{Gao_2019_CVPR}\cite{Gao_2019_ICCV} model to solve the VQA task, there are two common methods. One method is to use self-attention to encode each layer of text information and visual information, and then use the co-attention mechanism to realize information interaction between text modality and visual modality\cite{Gao_2019_CVPR}. Another~\cite{yu2019deep} method is to use self-attention mechanism to encode text information several times and learn high-level text information; Then high-level text information is used to interact with visual modality 
information by co-attention mechanism. These two methods are not considered enough because they do not reflect the impact of global text information formed by shallow and deep levels on multi-modal information interaction. So we use multi-layer global information of text to help multi-modal information interaction.

We first encode the text feature and the visual region feature three times using the self-attention mechanism. Considering the original text and visual information, so we have four layers of information in the text modality and the visual modality respectively. As shown in the following formula $rv$, $iv$, $jv$, $kv$ and $rw$, $iw$, $jw$, $kw$ represent the features of different layers; $Sa$ stands for self-attention encoding.

\vspace{-6pt}
\begin{eqnarray}
     rv  = v~~;iv{\rm{ }} =  {\rm{ S}}a\left( {rv} \right)~~;
     jv{\rm{ }} = {\rm{ S}}a\left( {iv} \right)~~;kv{\rm{ }} = {\rm{ S}}a\left( {jv} \right)\\
    rw  =  w~~;iw = Sa(rw)~~;jw = Sa(iw)~~;kw = Sa(jw)     
\end{eqnarray}

In order to reflect the overall information of each layer, we first average the text information of each layer. $\mathrm{Mean}$ represents the function of averaging.

\begin{equation}
xw_\mathrm{mean} = \mathrm{Mean}(xw),x \in \{r,i,j,k\}
\end{equation}

$rw_\mathrm{mean}$,$iw_\mathrm{mean}$,$jw_\mathrm{mean}$,$kw_\mathrm{mean}$ represents the overall information of each layer of text features, in order to reflect the correlation between the layers of information, we use RNN to process the mean information of each layer. $\mathrm{Multi}_{q_n}$ represents the overall information of the multi-layer text feature in a Quaternion Block.
\begin{equation}
\mathrm{Multi}_{q_n}= \mathrm{RNN}\left( {rw_\mathrm{mean},iw_\mathrm{mean},jw_\mathrm{mean},kw_\mathrm{mean}} \right)
\end{equation}
 We use the co-attention mechanism to achieve the fusion of visual information and text information, but unlike the previous method, we first add overall text features to each layer, so that the visual information of each layer is fused by the text information of the corresponding layer considering the overall feature of multi-layered text.

\begin{eqnarray}
        \mathrm{Multi}\_\mathrm{attention}\left( {query,key,value} \right)~~~~~~~\nonumber\\
        ~~= \mathrm{Concat}\left( {hea{d_1},...,hea{d_n}} \right)w \\
        head_i = \mathrm{Attention}\left( {Q,K,V} \right)
\end{eqnarray}
\vskip -5mm
\begin{eqnarray}
\mathrm{Attention}\left( {Q,K,V} \right) = & \nonumber \\
\mathrm{Softmax} \left( {\frac{{Q*K}}{{\sqrt {\dim }}}} \right)
*&{\!\!\!\!\!\!G_{\mathrm{quaternion - score - softmax} }} * V 
\end{eqnarray}
\vskip -5mm
\begin{eqnarray}
    xv_\mathrm{update}{\rm{ }} = {\rm{ }} \hspace {6.3cm} \nonumber\\
    \mathrm{Multi\_attention}{\rm{ }}\left( {xv,xw + {\rm{ }}\mathrm{Multi}_{q_n},xw + {\rm{ }}\mathrm{Multi}_{q_n}} \right)\\
    x \in \{r,i,j,k\} \nonumber 
\end{eqnarray}
\vskip -5mm
\begin{figure*}[t!]
 \centering
 \includegraphics[width=1.25\columnwidth]{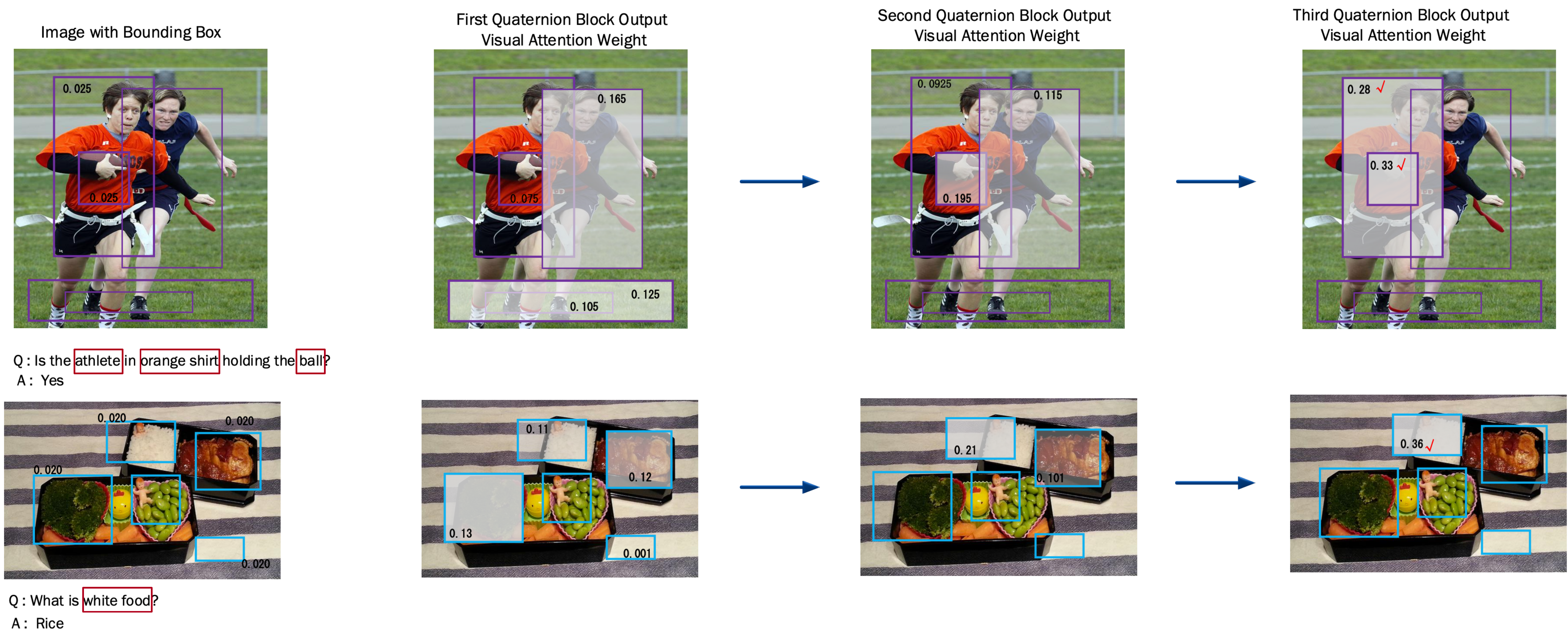}
 \vspace{-6pt}
 \caption{Visualization of the updating process of attention weights learned by our Quaternion Block Networks. If the weight of an object is higher, the color of that bounding box will be more ``gray".}
\label{fig:visualization}
\vspace{-6pt}
\end{figure*}

\subsection{Multi-layer feature relationship learning}\label{subsection:multi-ship}
For self-attention or co attention, only the same level of the attention map can be established. In order to establish the relationship between text modality and visual modal multi-layer features, we use the quaternion product method to learn this relationship and apply this relationship to Long interest in the attention. 
\begin{eqnarray}
    W{\rm{ }} & = & {\rm{ }}rw{\rm{ }} + iw\mathop I\limits^ \to  {\rm{ }} + jw\mathop J\limits^ \to  {\rm{ }} + kw\mathop K\limits^ \to \\
       V & = & {\rm{ }}rv{\rm{ }} + iv\mathop I\limits^ \to  {\rm{ }} + jv\mathop J\limits^ \to  {\rm{ }} + kv\mathop K\limits^ \to    
\end{eqnarray}
\vskip -0.5cm
\begin{equation} 
    \begin{array}{l}
    W \otimes V = \left( {rv * rw{\rm{  -  }}iv * iw{\rm{  - }}jv * jw - kv * kw} \right)\\
     + {\rm{ }}\left( {iv * rw{\rm{ }} + rv * iw{\rm{  -  }}kv * jw{\rm{ }} + jv * kw} \right)\mathop I\limits^ \to  \\
     + {\rm{ }}\left( {jv * rw{\rm{ }} + kv * iw{\rm{ }} + rv * jw{\rm{  - }}iv * kw} \right)\mathop J\limits^ \to  {\rm{ }}\\
     + {\rm{ }}\left( {kv * rw{\rm{  -  }}jv * iw{\rm{ }} + iv * jw{\rm{ }} + rv * kw} \right)\mathop K\limits^ \to  
    \end{array}
\end{equation}%

We separate the real and imaginary parts of the Hamilton Product's results; the real layer is represented by $r$, $i$ represents the first imaginary layer relationship, $j$ represents the second imaginary layer relationship, and $k$ represents the third imaginary layer relationship.  We apply these relationships to information updates between the image and text modalities~\cite{antol2015vqa}.
\begin{equation}
    \mathrm{quaternion\_score}_r{\rm{ }} = {\rm{ }}rv * rw{\rm{  -  }}iv * iw{\rm{  -  }}jv * jw{\rm{  -  }}kv * kw  
\end{equation}
\begin{equation}
    \mathrm{quaternion\_score}_i{\rm{ }} = {\rm{ }}iv * rw{\rm{ }} + rv * iw{\rm{  -  }}kv * jw{\rm{ }} + jv * kw   
\end{equation}
 \begin{equation}
    \mathrm{quaternion\_score}_j{\rm{ }} = {\rm{ }}jv * rw{\rm{ }} + kv * iw{\rm{ }} + rv * jw{\rm{  -  }}iv * kw   
\end{equation}
\begin{equation}
    \mathrm{quaternion\_score}_k{\rm{ }} = {\rm{ }}kv * rw{\rm{  -  }}jv * iw{\rm{ }} + iv * jw{\rm{ }} + rv * kw   
\end{equation}
\begin{eqnarray}
\mathrm{Quaternion\_softmax}_{\left( {r,i,j,k} \right)}{\rm{ }} \nonumber \hspace{3cm}\\
 = \mathrm{Softmax}\left( {\mathrm{quaternion\_score}_{\left( {r,i,j,k} \right)} } \right)    
\end{eqnarray}

\section{Experiments and Discussions}
\label{sec:print}

The evaluation results comparing with the conventional approaches are shown in Table 1. Table 2 shows the ablation study on the VQA 2.0 validation dataset to examine the performance by combining each module.  we tested our QBN on the test dataset by combining the VG dataset, the train and test dataset of VQA v2.0. All experiments use ADAM optimizer with 0.0001 learning rate for 13 epochs.

We mainly explore the effects of Multi-layer content learning, Multi-relationship content, dynamic sentence vector features effects on visual channel features, number of heads, number of stacks, and visual question-and-answer models in multi-layer content.

Table 2 shows the performance is improved after stacking the models, such as model BAN, DFAF, MCAN. This is because multi-level features are used to match the attention mechanism, which is helpful for improving the power of the model. The DFAF is a representative model in which the alignment levels are updated with each other. 
As shown in Fig.1, MCAN is a representative model that uses high-level information to match; Shown in Figure 3, our model first understands the multi-layer content of the text features in a quaternion-block, and obtains a representative multi-layer feature $\mathrm{Multi}_{qn}$, and then adds this feature, $\mathrm{Multi}_{qn}$, to the text information of each layer; When the image feature uses co-attention for information interaction, each layer feature of the text can be considered, and the corresponding layer feature is selected in the case of the character high-level feature. And in combination with the Quaternion-block gate, the model can achieve better results, because the $\mathrm{quaternion_{score}}$ in the quaternion-block gate can reflect the effective relationship between the features of other layers in the form of correlation coefficients, and the content of each layer. 
Importance, so that the image and the fine-grained head feature space in the text establish a more efficient relationship. 
The Scaled 2x2 region feature can bring richer visual information, which helps the performance.
	

\begin{table}[!htbp]
\footnotesize
\begin{tabular}{lccccc}
\toprule
\multirow{2}{*}{Model} &      \multicolumn{4}{c}{test-dev} & \multicolumn{1}{c}{test-std} \\
\cmidrule(lr){2-5} \cmidrule(lr){6-6}
 & Y/N  & No. & Other & All & All\\
\hline
Bottom-up~\cite{anderson2018bottom}&81.82&44.21&56.05&65.32&65.67\\
MFH~\cite{goyal2017making}& - & - & - & 66.12 & -\\
DCN~\cite{nguyen2018improved}&83.51&46.61&57.26&66.87&66.97\\
MFH+Bottom-Up~\cite{goyal2017making}&84.27&49.56&59.89&68.76& -\\
BAN+Glove~\cite{kim2018bilinear}&85.46&50.66&60.50&69.66& - \\
DFAF~\cite{Gao_2019_CVPR}&86.09&\underline{53.32}&60.49&70.22&70.34\\
DFAF-Bert~\cite{Gao_2019_CVPR}&86.73&52.92&\textbf{61.04}&70.59&70.81\\
MCAN~\cite{yu2019deep}&\underline{86.8}&53.26&60.72&\underline{70.63}&70.90\\
MLIN w/o Bert~\cite{Gao_2019_ICCV}&85.96&52.93&60.40&70.18&70.28\\
ViLBERT~\cite{lu2019vilbert}& - & - & - &70.55&\underline{70.92}\\
\hline
QBN (ours)&\textbf{87.12}&\textbf{53.48}&\underline{60.8}& \textbf{70.81}&\textbf{71.02} \\
\bottomrule
\end{tabular}
\caption{Experimental results on VQA 2.0 test-dev and test-std splits. In each sub-category, the best number is labeled as bold while the second largest number is underlined. Overall, our model achieves the best performance against previous strong baselines including pretraining-based approach “ViLBERT”.}
\vspace{-6pt}
\end{table}

\begin{table}[!htbp]
\footnotesize
\begin{tabular}{lcc}
\toprule
Component&Setting&Accuracy\\
\toprule
Bottom-up~\cite{anderson2018bottom}&Bottom-up&63.37\\
\hline 
Bilinear &BAN-1&65.36\\
Attention~\cite{kim2018bilinear} & BAN-12 & 66.04 \\
\hline
\multirow{2}{2cm}{DFAF~\cite{Gao_2019_CVPR}} & \multirow{2}{2.7cm}{\centering  DFAF-1\\ DFAF-8} & \multirow{2}{1cm}{\centering 66.21 \\ 66.66} \\ \\
\hline
MCAN~\cite{yu2019deep}&MCAN-9&67.16\\
\hline
\hline
\multirow{2}{3cm}{\# of Stacked Blocks} & \multirow{2}{2.7cm}{\centering QB-3 \\ QB-4} & \multirow{2}{1cm}{\centering 67.20 \\ 67.35} \\ \\
\hline
Embedding Dimension & 512 & \textbf{67.45}\\
\hline
\multirow{2}{2cm}{Region feature} & \multirow{2}{2.7cm}{\centering $1\times1\times2048$ \\ Scaled $2\times2\times2048$} & \multirow{2}{1cm}{\centering 67.35 \\ \textbf{67.45}} \\ \\
\hline
W/o Relationship Learning &  QB-4 & 67.28\\
\hline
W/o Content Learning & QB-4 & 67.25\\
\hline
\multirow{3}{2cm}{Parallel Heads} & \multirow{3}{2cm}{\centering 8 heads \\ \centering 12 heads \\ 16 heads} & \multirow{3}{1.5cm}{\centering  67.26 \\\centering  67.30 \\ 67.35} \\ \\ \\
\bottomrule
\end{tabular}
\caption{Ablation studies of Quaternion Block Networks on VQA v2.0 validation dataset. The N in QB-N represents the number of a Quaternion block.}
\label{tab:valTable}
\vspace{-6pt}
\end{table}
\section{Conclusion}
Our proposed Multi-Layer Content Interaction can be a fundamental approach to capture high-level interaction between different modalities. In this paper, the performance on Visual Question Answering can even surpass models with BERT pre-training. In the future, we will test our model on more diverse applications like visual grounding and vision language pre-training.\\ 
\textbf{Acknowledgement}
This work was supported in part by National key research and development program of China(2016QY01W0200), The Foundation for Innovative Research Groups of the National Natural Science Foundation of China(Grant No. 61921003).

\label{sec:page}

\vfill\pagebreak

{\small
\bibliographystyle{IEEEbib}
\bibliography{refs}

\begin{thebibliography}{10}

\bibitem{hori2017attention}
Chiori Hori, Takaaki Hori, Teng-Yok Lee, Ziming Zhang, Bret Harsham, John~R
  Hershey, Tim~K Marks, and Kazuhiko Sumi,
\newblock ``Attention-based multimodal fusion for video description,''
\newblock in {\em ICCV}, 2017.

\bibitem{gao2018question}
Peng Gao, Hongsheng Li, Shuang Li, Pan Lu, Yikang Li, Steven~CH Hoi, and
  Xiaogang Wang,
\newblock ``Question-guided hybrid convolution for visual question answering,''
\newblock in {\em Proceedings of the European Conference on Computer Vision
  (ECCV)}, 2018, pp. 469--485.

\bibitem{hori2019end}
Chiori Hori, Huda Alamri, Jue Wang, Gordon Wichern, Takaaki Hori, Anoop
  Cherian, Tim~K Marks, Vincent Cartillier, Raphael~Gontijo Lopes, Abhishek
  Das, et~al.,
\newblock ``End-to-end audio visual scene-aware dialog using multimodal
  attention-based video features,''
\newblock in {\em ICASSP 2019-2019 IEEE International Conference on Acoustics,
  Speech and Signal Processing (ICASSP)}. IEEE, 2019, pp. 2352--2356.

\bibitem{Gao_2019_CVPR}
Peng Gao, Zhengkai Jiang, Haoxuan You, Pan Lu, Steven C.~H. Hoi, Xiaogang Wang,
  and Hongsheng Li,
\newblock ``Dynamic fusion with intra- and inter-modality attention flow for
  visual question answering,''
\newblock in {\em The IEEE Conference on Computer Vision and Pattern
  Recognition (CVPR)}, June 2019.

\bibitem{yu2019deep}
Zhou Yu, Jun Yu, Yuhao Cui, Dacheng Tao, and Qi~Tian,
\newblock ``Deep modular co-attention networks for visual question answering,''
\newblock in {\em CVPR}, 2019.

\bibitem{goyal2017making}
Yash Goyal, Tejas Khot, Douglas Summers-Stay, Dhruv Batra, and Devi Parikh,
\newblock ``Making the v in vqa matter: Elevating the role of image
  understanding in visual question answering,''
\newblock in {\em CVPR}, 2017.

\bibitem{simonyan2014very}
Karen Simonyan and Andrew Zisserman,
\newblock ``Very deep convolutional networks for large-scale image
  recognition,''
\newblock {\em ICLR}, 2015.

\bibitem{he2016deep}
Kaiming He, Xiangyu Zhang, Shaoqing Ren, and Jian Sun,
\newblock ``Deep residual learning for image recognition,''
\newblock in {\em CVPR}, 2016.

\bibitem{hu2018squeeze}
Jie Hu, Li~Shen, and Gang Sun,
\newblock ``Squeeze-and-excitation networks,''
\newblock in {\em CVPR}, 2018.

\bibitem{huang2017densely}
Gao Huang, Zhuang Liu, Laurens Van Der~Maaten, and Kilian~Q Weinberger,
\newblock ``Densely connected convolutional networks,''
\newblock in {\em CVPR}, 2017.

\bibitem{mikolov2013distributed}
Tomas Mikolov, Ilya Sutskever, Kai Chen, Greg~S Corrado, and Jeff Dean,
\newblock ``Distributed representations of words and phrases and their
  compositionality,''
\newblock in {\em NIPS}, 2013.

\bibitem{pennington2014glove}
Jeffrey Pennington, Richard Socher, and Christopher Manning,
\newblock ``Glove: Global vectors for word representation,''
\newblock in {\em EMNLP}, 2014.

\bibitem{kiros2015skip}
Ryan Kiros, Yukun Zhu, Ruslan~R Salakhutdinov, Richard Zemel, Raquel Urtasun,
  Antonio Torralba, and Sanja Fidler,
\newblock ``Skip-thought vectors,''
\newblock in {\em NIPS}, 2015.

\bibitem{peters2018deep}
Matthew~E Peters, Mark Neumann, Mohit Iyyer, Matt Gardner, Christopher Clark,
  Kenton Lee, and Luke Zettlemoyer,
\newblock ``Deep contextualized word representations,''
\newblock {\em NAACL}, 2018.

\bibitem{devlin2018bert}
Jacob Devlin, Ming-Wei Chang, Kenton Lee, and Kristina Toutanova,
\newblock ``Bert: Pre-training of deep bidirectional transformers for language
  understanding,''
\newblock {\em NAACL}, 2019.

\bibitem{mnih2014recurrent}
Volodymyr Mnih, Nicolas Heess, Alex Graves, et~al.,
\newblock ``Recurrent models of visual attention,''
\newblock in {\em NIPS}, 2014.

\bibitem{bahdanau2014neural}
Dzmitry Bahdanau, Kyunghyun Cho, and Yoshua Bengio,
\newblock ``Neural machine translation by jointly learning to align and
  translate,''
\newblock {\em ICLR}, 2015.

\bibitem{vaswani2017attention}
Ashish Vaswani, Noam Shazeer, Niki Parmar, Jakob Uszkoreit, Llion Jones,
  Aidan~N Gomez, {\L}ukasz Kaiser, and Illia Polosukhin,
\newblock ``Attention is all you need,''
\newblock in {\em NIPS}, 2017.

\bibitem{Gao_2019_ICCV}
Peng Gao, Haoxuan You, Zhanpeng Zhang, Xiaogang Wang, and Hongsheng Li,
\newblock ``Multi-modality latent interaction network for visual question
  answering,''
\newblock in {\em The IEEE International Conference on Computer Vision (ICCV)},
  October 2019.

\bibitem{pavllo2018quaternet}
Dario Pavllo, David Grangier, and Michael Auli,
\newblock ``Quaternet: A quaternion-based recurrent model for human motion,''
\newblock {\em BMVC}, 2018.

\bibitem{parcollet2018quaternion}
Titouan Parcollet, Ying Zhang, Mohamed Morchid, Chiheb Trabelsi, Georges
  Linar{\`e}s, Renato De~Mori, and Yoshua Bengio,
\newblock ``Quaternion convolutional neural networks for end-to-end automatic
  speech recognition,''
\newblock {\em INTERSPEECH}, 2018.

\bibitem{tay2019lightweight}
Yi~Tay, Aston Zhang, Luu~Anh Tuan, Jinfeng Rao, Shuai Zhang, Shuohang Wang, Jie
  Fu, and Siu~Cheung Hui,
\newblock ``Lightweight and efficient neural natural language processing with
  quaternion networks,''
\newblock {\em ACL}, 2019.

\bibitem{ren2015faster}
Shaoqing Ren, Kaiming He, Ross Girshick, and Jian Sun,
\newblock ``Faster r-cnn: Towards real-time object detection with region
  proposal networks,''
\newblock in {\em NIPS}, 2015.

\bibitem{krishna2017visual}
Ranjay Krishna, Yuke Zhu, Oliver Groth, Justin Johnson, Kenji Hata, Joshua
  Kravitz, Stephanie Chen, Yannis Kalantidis, Li-Jia Li, David~A Shamma,
  et~al.,
\newblock ``Visual genome: Connecting language and vision using crowdsourced
  dense image annotations,''
\newblock {\em IJCV}, 2017.

\bibitem{anderson2018bottom}
Peter Anderson, Xiaodong He, Chris Buehler, Damien Teney, Mark Johnson, Stephen
  Gould, and Lei ZBotthang,
\newblock ``Bottom-up and top-down attention for image captioning and visual
  question answering,''
\newblock in {\em CVPR}, 2018.

\bibitem{antol2015vqa}
Stanislaw Antol, Aishwarya Agrawal, Jiasen Lu, Margaret Mitchell, Dhruv Batra,
  C~Lawrence~Zitnick, and Devi Parikh,
\newblock ``Vqa: Visual question answering,''
\newblock in {\em ICCV}, 2015.

\bibitem{nguyen2018improved}
Duy-Kien Nguyen and Takayuki Okatani,
\newblock ``Improved fusion of visual and language representations by dense
  symmetric co-attention for visual question answering,''
\newblock in {\em CVPR}, 2018.

\bibitem{kim2018bilinear}
Jin-Hwa Kim, Jaehyun Jun, and Byoung-Tak Zhang,
\newblock ``Bilinear attention networks,''
\newblock in {\em NIPS}, 2018.

\bibitem{lu2019vilbert}
Jiasen Lu, Dhruv Batra, Devi Parikh, and Stefan Lee,
\newblock ``Vilbert: Pretraining task-agnostic visiolinguistic representations
  for vision-and-language tasks,''
\newblock {\em arXiv}, 2019.

\end{thebibliography}
}

\end{document}